# Cross-regional oil palm tree counting and detection via multi-level attention domain adaptation network

Juepeng Zheng [a, b], Haohuan Fu [a, b], Weijia Li [a, b, c, *], Wenzhao Wu [a, b], Yi Zhao [a, b], Runmin Dong [a, b], Le Yu [a, b]

[a] Ministry of Education Key Laboratory for Earth System Modeling, Department of Earth System Science, Tsinghua University, Beijing 100084, China

[b] Joint Center for Global Change Studies, Beijing 100875, China

[c] CUHK-SenseTime Joint Lab, The Chinese University of Hong Kong, Hong Kong, China

* Corresponding author: weijiali@cuhk.edu.hk

Abstract

Providing an accurate evaluation of palm tree plantation in a large region can bring meaningful impacts in both economic and ecological aspects. However, the enormous spatial scale and the variety of geological features across regions has made it a grand challenge with limited solutions based on manual human monitoring efforts. Although deep learning based algorithms have demonstrated potential in forming an automated approach in recent years, the labelling efforts needed for covering different features in different regions largely constrain its effectiveness in large-scale problems. In this paper, we propose a novel domain adaptive oil palm tree detection method, i.e., a Multi-level Attention Domain Adaptation Network (MADAN) to reap cross-regional oil palm tree counting and detection. MADAN consists of 4 procedures: First, we adopted a batch-instance normalization network (BIN) based feature extractor for improving the generalization ability of the model, integrating batch normalization and instance normalization. Second, we embedded a multi-level attention mechanism (MLA) into our architecture for enhancing the transferability, including a feature level attention and an entropy level attention. Then we designed a minimum entropy regularization (MER) to increase the confidence of the classifier predictions through assigning the entropy level attention value to the entropy penalty. Finally, we employed a sliding window-based prediction and an IOU based post-processing approach to attain the final detection results. We conducted comprehensive ablation experiments using three different satellite images of large-scale oil palm plantation area with six transfer tasks.

MADAN improves the detection accuracy by 14.98% in terms of average F1-score compared with the Baseline method (without DA), and performs 3.55%-14.49% better than existing domain adaptation methods. Experimental results demonstrate the great potential of our MADAN for large-scale and cross-regional oil palm tree counting and detection, guaranteeing a high detection accuracy as well as saving the manual annotation efforts.

Keywords: Oil palm tree detection; Attention mechanism; Domain adaptation; Deep learning; Adversarial neural networks

## 1. Introduction

Oil palm, an economic perennial crop mostly cultivated across the Southeast Asia, is an important source of edible oils and fats (Rhys et al., 2018). It is also used for producing oleo chemicals, which are the main ingredients of personal care products, cosmetics and cleaning products. The demand for palm oil is increasing and the production is estimated to reach 72 million tons by 2019 with Malaysia and Indonesia as leading producers, accounting for more than 80% of the global production and dominating the international trade (Koh, & Wilcove, 2007; Cheng et al., 2016; Senawi et al., 2019; Truckell et al., 2019). Recently, oil palm has attracted lots of attention from governments and researchers, because it plays an essential role in maintaining carbon balance and possessing high economic value. In addition, the expansion of oil palm plantation area is condemned by the environmental protectors due to threatening the survival of native species, destroying the tropical rain forest and reducing the biodiversity (Busch et al., 2015; Cheng et al., 2017; Cheng et al., 2018; Carlson et al., 2018; Quezada et al., 2019). As a result, counting and detecting oil palm trees from high-resolution remotely sensed images is a significant work for better management, efficient fertilization and irrigation of oil palm plantation. Owing to the development of machine learning and deep learning, many tree crown detection methods have been proposed with satisfying performance. However, most of them focus on detecting tree crowns in a single study region using single-source remote sensing imagery (Daliakopoulos et al., 2009; Dalponte et al., 2014; Li et al., 2017; Li et al., 2019a; Wang et al., 2019b; Feng et al., 2019).

Large-scale and cross-regional oil palm tree investigation is a pivotal research issue. Nowadays, the affluent remote sensing images and rapid development of deep learning algorithms bring new opportunity to large-scale and cross-regional oil palm detection. However, large-scale tree counting and detection may be confronted with remote sensing images with diverse acquisition conditions, like different sensors, seasons and environments, resulting in different distribution and domain shifts among images. For example, as is shown in Figure 1, Image A and Image B are two different satellite images. Here we assume Image A as source domain that has enough labels while Image B as target domain that has no label for training. We can easily observe the obvious discrepancy between two images in terms of the histogram of each class, resulting from the difference in sensors, acquisition dates and locations. That is, even if we have an outstanding oil palm detection accuracy in a particular scenario, like Image A, when it is directly applied to a new data set without any labels, like Image B, the performance of the detector may drop dramatically.

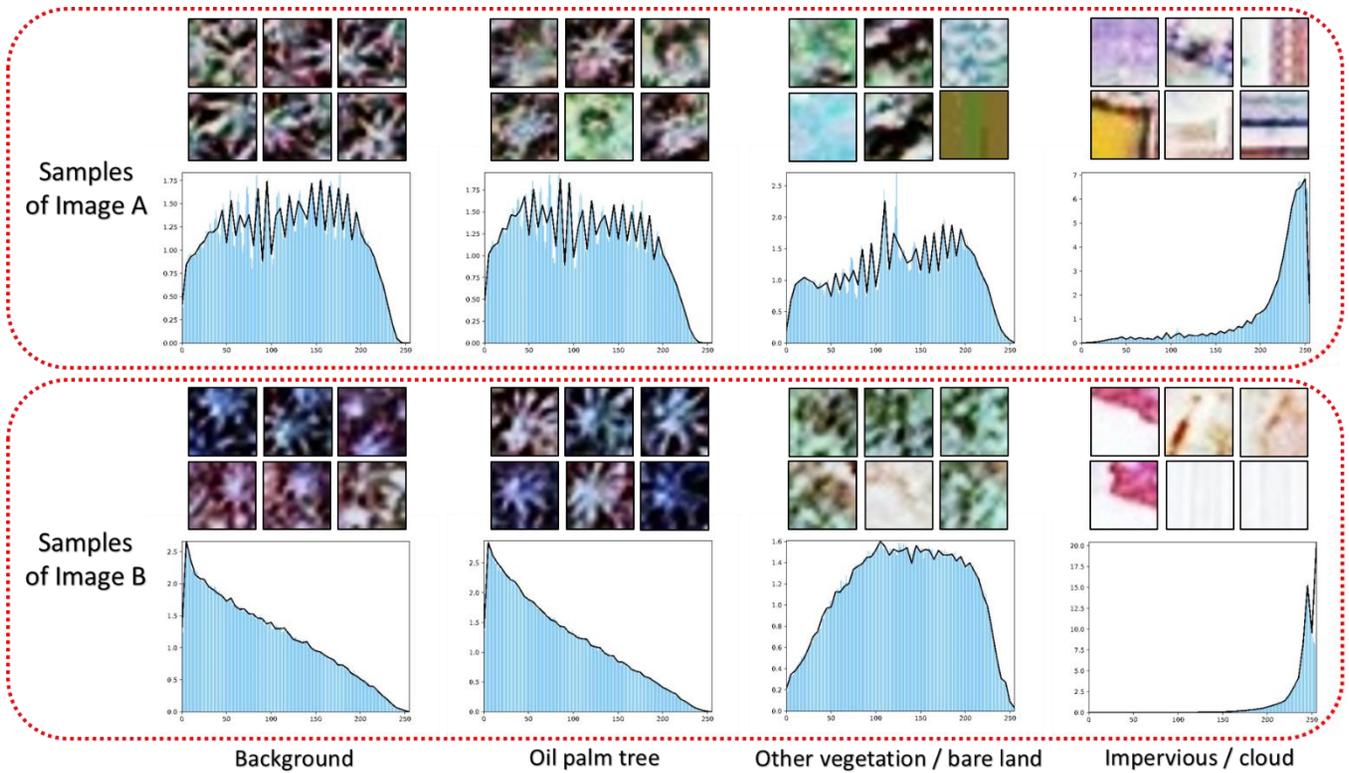

Figure 1. The spectral distribution of different classes in Image A and Image B. The histograms denote the mean histogram of all training samples in one category for Image A and Image B, respectively. We can easily observe the obvious discrepancy between two images in terms of the histogram of each class.

Fortunately, domain adaptation (DA) methods can help adapting the model to new data domains without leveraging a large quantity of costly labels, which has attracted lots of attention over the past decades. According to the availability of labelled samples from the target domain, DA can be classified into unsupervised DA (UDA), semi-supervised DA (SSDA) and supervised DA (Wang & Deng, 2018). UDA directly aims at improving the generalization capability of the model between the source domain and the target domain, without using any labels from the target domain (Ghifary et al., 2014; Ganin et al., 2016). SSDA allows the model to learn the information from the target domain, based on slight labelled data in the target domain and sufficient labeled data in the source domain (Kumar et al., 2010; Donahue et al., 2013). Supervised DA uses a small number of labelled data in the target domain that are usually not sufficient for tasks (Chopra et al., 2013; Tzeng et al., 2015). In this paper, we concentrate on UDA method, a promising type of method in transfer learning, which only requires labels in the source domain and completely label-free in the target domain.

As traditional deep learning methods focus on grasping the texture patterns in different images (LeCun et al., 2015), the performance of trained network in one set of satellite images would degrade significantly when moving to images that are taken in a different region or from a different source. However, thus far, the use of domain adaptation techniques for cross-regional remote sensing image processing is still at a considerably earlier stage with the following limitations (Tuia et al., 2016). First, most of these studies focus on land cover and land use classification, hyperspectral images classification, and scene classification issues (Bruzzone, & Persello, 2009; Matasci et al., 2012). Second, these studies usually focus only on decreasing the distribution discrepancy between the source domain dataset and the target domain dataset (or a small-scale in a local area) instead of a large-scale target region (Zhu et al., 2019; Ma et al., 2019). Moreover, they may not take full advantages of the existing annotations and exploit the transferability of different samples in the source domain, resulting in a prominent gap of accuracy between the source and the target domain.

In this paper, we propose a novel UDA based oil palm tree counting and detection algorithm, i.e., a Multi-Attention

Domain Adaptation Network (MADAN), to improve the oil palm tree detection performance across different remotely sensed images acquired from different sensors, regions and dates, without using labeled samples in the target region. Our MADAN is proposed for enhancing both the generalization capacity and the transferability of our model. Our codes and datasets are available on https://github.com/rs-dl/MADAN. The major contributions of our work are as follows:

(1) We propose an adaptive object detector named MADAN for oil palm tree counting and detection across different satellite images, which is the first work for large-scale domain adaptive tree crown detection using multi-source and multi-temporal remote sensing images.

(2) We integrate batch normalization and instance normalization as Batch-Instance Normalization (BIN) block. BIN block is embedded into our neural network backbone, improving both its generalization capacity and classification performance.

(3) We design two types of transferrable attention mechanisms for our DA part, i.e., the feature level attention mechanism and the entropy level attention mechanism. Both of them are generated by the domain discriminator through adversarial learning that effectively exploits the transferability of the sample datasets between the source and the target domains. The higher transferability an image has, the higher attention value it achieves. That is, the image with more transferrable context will have a higher weight in the neural networks.

The remainder of this paper is organized as follows. We introduce the related work about tree crown detection and DA in next section. Following that, we introduce our proposed MADAN in detail in Section 3, and present our large-scale study area and dataset in Section 4. We analyze and compare the oil palm tree counting and detection results of our proposed MADAN and other DA approaches in Section 5, followed by comprehensive ablation experiments in Section 6. At last, we summarize our paper and present our future works in Section 7.

## 2. Related work

### 2.1 Tree crown detection

The previous work related to tree crown detection can be classified into classical image processing methods, traditional machine learning methods, and deep learning methods. Classical image processing methods usually include image binarization, local maximum filter and image segmentation, etc. (Daliakopoulos et al., 2009; Wulder et al., 2000; Chemura et al., 2015). Although these methods do not require labels, complex scenarios such as overlapping tree crowns may cause deterioration of detection results. As for traditional machine learning methods, many algorithms have been applied to tree crown detection, including random forest, support vector machine (SVM), artificial neural network, etc. (Pu, & Landry, 2012; Hung et al., 2012; Dalponte et al., 2014, Wang et al., 2019b). For instance, Wang et al. (2019b) automatically detect oil palms in Malaysia for Unmanned Aerial Vehicle (UAV) images using the histogram of oriented gradient (HOG) and SVM classifier, which obtained the overall accuracy of over 94%. Traditional machine learning methods make great progress compared with classical image processing methods, but most of them require sophisticated techniques or very-high-resolution UAV images. Following the achievement of AlexNet (Krizhevsky et al., 2012), many deep learning algorithms have been developed and successfully adopted to lots of remote sensing tasks since 2014 (Wu et al., 2019; Li et al., 2019b; Ienco et al., 2019; Dong et al., 2019). In 2016, Li et al. (2016b) applied deep learning-based method to tree crown detection. After that, they proposed a two-stage CNN that achieved a higher average F1-score of 92.80%, exceeding single-stage CNN and other traditional machine learning based methods (Li et al., 2019a). Moreover, Mubin et al. (2019) and Neupane et al. (2019) utilized sliding window-based approach combined with deep learning to detect oil palm trees and banana plants, respectively. In summary, most of recent studies adopt a machine learning or deep learning-based classifier combined with a sliding window-based method to detect tree crowns from satellite images.

Deep learning, which is known for its remarkable capacity of feature extraction, requires a large number of labeled samples. The aforementioned methods were only applied to detecting tree crowns in a particular region, and both the training and test images were photographed in the same condition. Existing studies have not explored the generalization and transferability of their models and simply focus on the accuracy of local regions, assuming that they were strong enough in areas without training samples.

**2.2 Domain adaptation**

DA belongs to transfer learning, and it has been widely used for image classification. In general, we can classify DA into three cases: discrepancy-based DA, adversarial-based DA and reconstruction-based DA according to Csurka (2017). The first case, discrepancy-based DA, such as Kullback-Leibler (KL) divergence, maximum mean discrepancy (MMD) and correlation alignment (CORAL), etc. (Ghifary et al., 2014; Tzeng et al., 2014; Long et al., 2015; Zhuang et al., 2015; You et al., 2019; Wang et al., 2019c), assumes that fine-tuning the deep network model with labeled or unlabeled target data can diminish the shift between the two domains. According to Wang & Deng (2018), class criterion, statistic criterion, architecture criterion and geometric criterion are four major techniques for performing fine-tuning. As for adversarial-based DA case, scholars assign a discriminator to classify whether an image is derived from the source domain or the target domain, and try to train the discriminator not to distinguish the two domains well by an adversarial objective, mapping the target images to the same space (Ganin, & Lempitsky, 2015; Ganin et al., 2016; Wang et al., 2019a; Chen et al., 2019; Wang et al., 2020). The main idea in reconstruction-based DA is diminishing the differences between the original and reconstructed images via generative adversarial network (GAN) discriminator (Ghifary et al., 2016; Kim et al., 2017).

DA has been exploited in the remote sensing community to cope with multi-temporal and multi-source satellite images, where differences in atmospheric illuminations and ground conditions can easily ruin the adaptation of a model (Bruzzone, & Persello, 2009; Volpi et al., 2015; Matasci et al., 2015; Samat et al., 2016; Yan et al., 2018b; Yan et al., 2019; Zhu et al., 2019; Ma et al., 2019). In 2009, Bruzzone & Persello (2009) proposed a feature selection method accomplished by a multi-objective criterion function to improve the discrimination in hyperspectral image classification. After that, they presented an approach to iteratively label and add samples, as well as remove the samples in the source domain that do not fit with the target domain. That is, they use active learning (AL) to address DA problems. Matasci et al. (2015) analyzed the effectiveness of TCA in multi- and hyperspectral image classification, and explored its unsupervised and semi-supervised implementation. Samat et al. (2016) used GFK based SVM for hyperspectral image classification to solve the different distributions between training and validation datasets. The literatures mentioned above applied traditional DA methods to remote sensing image classification. For deep learning based DA approaches, Zhu et al. (2019) proposed a semi-supervised adversarial learning domain adaptation framework for scene classification, reaching the overall accuracy of over 93% in different temporal aerial images. Ma et al. (2019) presented a deep DA method for hyperspectral image classification based on a domain alignment module, a task allocation module, and a DA module.

Existing DA methods mainly concentrate on the classification task. Hence, almost all of these studies utilized DA methods for land cover and land use classification, hyperspectral images classification, and scene classification. On the contrary, the study of domain adaptive object detection and semantic segmentation is still at a considerably earlier stage (Yan et al., 2018a; Benjdira et al., 2019; Koga et al., 2020). In semantic segmentation, Benjdira et al. (2019) used GANs to reduce the domain shift of aerial images, improving the average segmentation accuracy from 14% to 61%. As for object detection, to the best of our knowledge, Koga et al. (2020) firstly applied CORAL and adversarial DA to vehicle detection from satellite images so far. However, although they improve the result of vehicle detection from 66.3% to 76.8%, there still exist a nonnegligible gap between the source and the target domain, of which the accuracy is almost 10% lower than the upper bound (obtained through directly training on the target dataset).

Based on the above analysis, in this paper, we proposed MADAN, a domain adaptive oil palm tree counting and detection algorithm. MADAN improves the capacity of generalization and transferability for cross-regional oil palm tree detection from multi-source and large-scale remote sensing images. It is the first work for large-scale domain adaptive tree crown detection, achieving a high detection accuracy and reducing the manual annotation efforts.

## 3. MADAN

### 3.1 Overview of our proposed method

In this paper, we concentrate on unsupervised domain adaptive oil palm detection across two different remote sensing images, which consists of an annotated source domain dataset $D_S = \{(x_i^S, y_i^S)\}_{i=1}^{n_S}$ in the source region ($R_S$) and an unlabeled target domain dataset $D_T = \{x_i^T\}_{i=1}^{n_T}$ in the target region ($R_T$), where $x_i$ is an example and $y_i$ is the corresponding label; $n_S$ and $n_T$ are the quantity of samples in the source and target dataset, respectively. Notably, $D_S$ is collected from $R_S$ and $D_T$ is collected from $R_T$. DA problems usually focus only on decreasing the distribution discrepancy between $D_S$ and $D_T$, while in remote sensing domain, the key goal is to improve the prediction results not only in $D_T$, but also in $R_S$ and $R_T$. As a result, we added Batch-Instance Normalization (BIN) blocks into the deep network to improve its generalization ability. For example, in Figure 2, we assume Image A as the source region ($R_S$) and Image B as the target region ($R_T$). The collected datasets in Image A and Image B are $D_S$ and $D_T$, respectively. Our goal is to boost the classification accuracy of $D_T$ and improve the detection result of $R_T$ (i.e. Image B).

In our framework, we manually annotated samples in the source domain, and randomly selected a suitable amount of images (without manually annotated labels) in the target domain. Figure 2 shows the framework of MADAN, including a BIN-based feature extractor, an attention-based adversarial learning with minimum entropy regularization, and an IOU based post-processing. We summarize the four major procedures of MADAN as follows.

(1) A BIN-based feature extraction for improving the generalization capacity. Both the labelled source data and the unlabeled target data are used as the input of the deep neural network. Here, we use 5 convolutional layers and 1 pooling layer in our feature extractor as the size of an input image is only 17×17 pixels. Following each convolutional layer, we integrate a batch normalization layer and an instance normalization layer, followed by an activation layer.

(2) An adversarial learning based multi-level attention mechanism for improving the transferability. We propose a feature level attention and an entropy level attention, which are generated by shallow feature and deep feature based adversarial discriminators, respectively. The feature level attention is assigned to the feature map and the entropy level attention is assigned to the entropy penalty. Meanwhile, besides the deep feature based domain loss, we add a shallow feature based domain loss to avoid information loss because of pooling layer.

(3) A minimum entropy regularization for improving the prediction confidence. Besides label prediction loss in the source domain, we add a minimum entropy regularization with an entropy level attention value to enhance the positive transfer for each image.

(4) Sliding window based reference and IOU based post-processing. After training the whole model, we partition the test images with overlaps, and predict the type of each sample in the test images. At last, we adopt IOU based metric to merge the detected oil palm trees that are very close, then we can get the final detection results in test images.

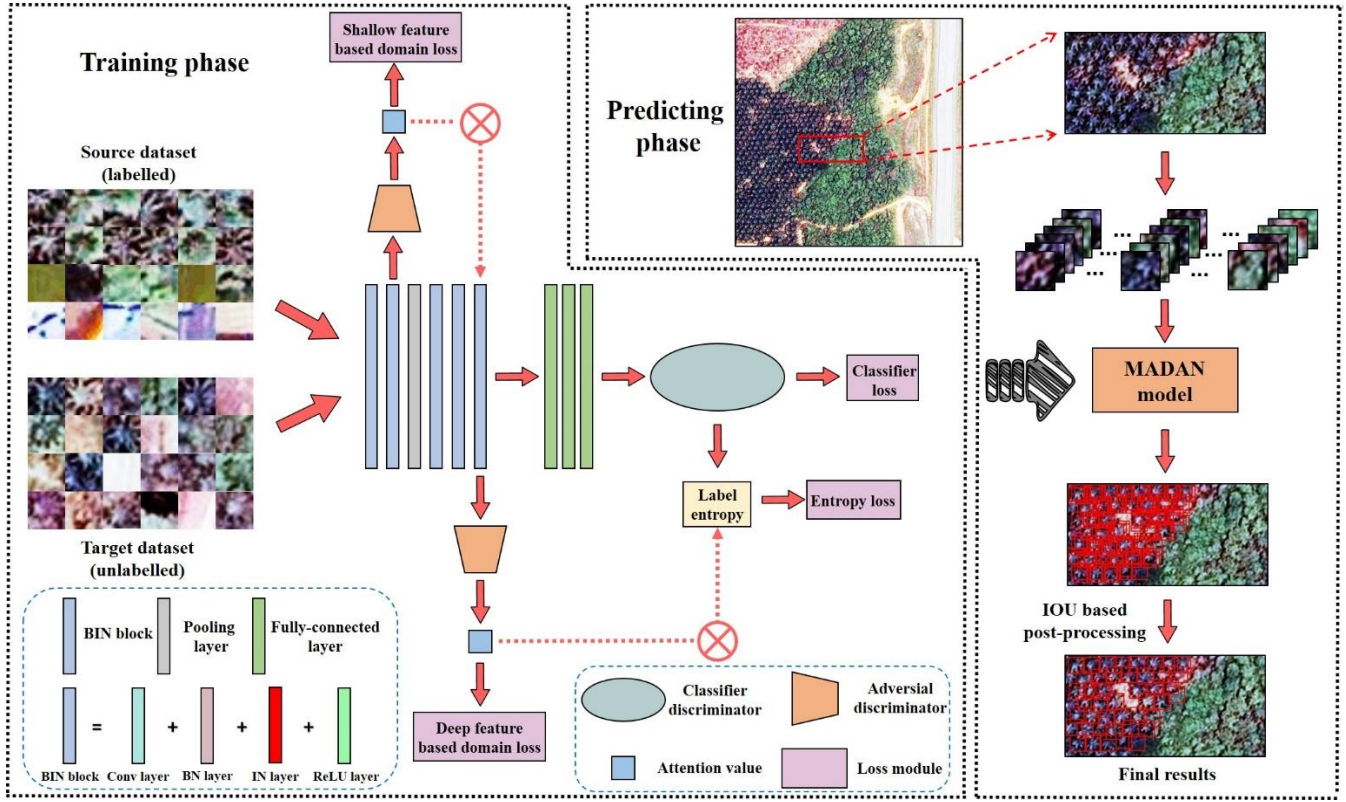

Figure 2. The overall framework of our proposed MADAN.

## 3.2 BIN based feature extractor

AlexNet was proposed in 2012 and received great success in the academic community. It consists of 5 convolutional layers, 3 max-pooling layers, and 3 fully-connected layers (Krizhevsky et al., 2012). Our network is similar to AlexNet, consisting of 5 convolutional layers and 3 fully-connected layers, while only 1 max-pooling layer is used as the size of our input image is only 17 × 17 pixels. Inspired by IBN-Net (Pan et al., 2018), we replace each convolutional layer of AlexNet with a BIN block, which is proposed to strengthen the generalization of our model.

The first two layers of our BIN block are the convolutional layer and the batch normalization (BN) layer (Ioffe et al., 2015). Although BN can effectively accelerate the convergence of the model, it makes CNNs vulnerable to appearance transforms such as the discrepancy between different satellite images for oil palm detection. So we add an instance normalization (IN) layer to eliminate the individual contrast as well as keep appearance and visual invariance, which was exploited in style transfer at first (Ulyanov et al., 2016). As illustrated in the left-bottom of Figure 2, each BIN block contains a convolutional layer, a BN layer, an IN layer and a ReLU layer. Our BIN based feature extractor helps to enhance the generalization of our architecture. That is, we only use labeled source domain dataset and assume that the target domain is completely "unseen" before.

### 3.3 Adversarial learning based multi-level attention mechanism (MLA)

Adversarial learning based multi-level attention mechanism is designed for improving the transferability of our model. The adversarial learning has been successfully used in previous domain adaptation studies (Ganin, & Lempitsky, 2015; Ganin et al., 2016; Chen et al., 2019). Attention mechanism has attained significant effect in improving the performance of deep learning (Chen et al., 2016; Wang et al., 2017). Transferable attention enables the model to pay more attention to transferable information of an image across domains, through assigning different weights for each pixel of an image. Wang et al. (2019a) applied attention mechanism to domain adaptation, which not only enables the model to pay attention to an image from the source domain, but also connect this attention to an image of interest from the target domain. In our proposed approach, we designed a multi-level attention mechanism through adversarial learning, including a feature level attention and an entropy level attention. In our proposed attention mechanism, the higher transferability an image has, the higher attention value it achieves. That is, the image with more transferrable context will have a higher weight in the neural networks. Details of the feature level attention and entropy level attention are introduced as follows.

### 3.3.1 Shallow feature based domain loss and feature level attention

As introduced in Section 4.2, we obtain the features from source and target datasets using a BIN based feature extractor, which are denoted by $F^S$ and $F^T$, respectively. However, above features are generated by a pooling layer, which may lose some effective information in shallow features. As a consequence, our feature level attention is generated by the features before the pooling layer, which are denoted by $F'^S$ and $F'^T$. As mentioned above, images do not perform equally well for transferring across domains, and some input images are more transferable than others. For example, the images from the target dataset that are significantly dissimilar in the feature space across domains may have a negative effect for transferability, and thus they may confuse the classifier. Accordingly, we apply a domain discriminator to obtain the feature level attention via adversarial learning, and the loss function of the shallow feature based domain discriminator ($L_S$) can be formulated as:

$$L_S^S = \frac{1}{n_S} \sum_{f_i'^S \in F'^S} L_d(G_d(f'^S_i), d_i^S) \tag{1}$$

$$L_S^T = \frac{1}{n_T} \sum_{f_i'^T \in F'^T} L_d(G_d(f'^T_i), d_i^T) \tag{2}$$

$$L_S = L_S^S + L_S^T \tag{3}$$

where $F'^S = \{f_i'^S\}_{i=1}^{n_S}$ and $F'^T = \{f_i'^T\}_{i=1}^{n_T}$. $G_d$ is the shallow feature based domain discriminator and $L_d$ is the cross-entropy loss of $G_d$. $d_i$ is equal to 1 for the source domain dataset and 0 for the target domain dataset. For more explanation, the output of $G_d(f_i)$ is the probability ($d_i^F$) of the feature map in image $i$ belonging to the source domain. When $d_i^F$ is larger than 0.5, it denotes that the feature map belongs to the source domain, and when the probability is lower than 0.5, it represents that it belongs to the target domain. The goal of our feature level attention is to find the images that have a good capacity of transferability between source and target domains. So in order to pay more attention to images with higher transferability, we use information entropy, also called Shannon entropy, to describe uncertainty, which is defined as $E(p) = -\sum_d p_d \cdot \log(p_d)$, where $p_{d=0}$ means the probability of the image belonging to the target domain while $p_{d=1}$ represents the probability of the image belonging to the source domain. According to the information theory, the larger the entropy is, the more information the probability has and the better transferability the image has. In other words, if $d_i^F$ is approaching 0.5, our network is harder to identify whether image $i$ is belonging to the source or the target domain, and thus the image is more transferrable. So the final feature level attention value ($v_i^F$) for each feature map can be calculated as:

$$v_i^F = 1 + E(d_i^F) \quad (4)$$

In this way, we can effectively quantify the transferability of the image. Then we are supposed to tell the network which feature maps are fitting for our cross-regional oil palm tree detection, and which feature maps may have negative effect to some extent. Accordingly, inspired by Wang et al., 2017, we add a connection between the feature map and the feature level attention value, and finally transformed the feature map according to formula (13):

$$h_i = f_i \cdot (1 + v_i^F) \quad (5)$$

where $f_i$ is the feature in the last convolutional layer and $h_i$ is the new feature map containing the information of transferability, in which the features with better transferability are weighted by a higher feature level attention value.

### 3.3.2 Deep feature based domain loss and entropy level attention

In Section 4.3.1, we obtain the deep features from the 5[th] BIN block with feature level attention. We can calculate the deep feature based domain loss ($L_D$) according to following formulas:

$$L_D^S = \frac{1}{n_S} \sum_{h_i^S \in H^S} L_d(G_d(h_i^S), d_i^S) \quad (6)$$

$$L_D^T = \frac{1}{n_T} \sum_{h_i^T \in H^T} L_d(G_d(h_i^T), d_i^T) \tag{7}$$

$$L_D = L_D^S + L_D^T \tag{8}$$

where $L_D^S$ and $L_D^T$ denote the source and the target domain loss based on deep features with feature level attention. $L_d$, $G_d$, $d_i^S$ and $d_i^T$ are the same as those used in shallow feature based domain loss. In this way, our domain loss comprises the shallow feature based domain loss and the deep feature based domain loss, which comprehensively consider the transferability of the feature maps from the last and the 2nd BIN block. Similar to the feature level attention, we define an entropy level attention that is assigned to the entropy loss, which is introduced in Section 4.4. Images that are not transferable in our domain adaptive method may have a negative effect on entropy loss, thus our entropy level attention value ($v_i^E$) can be defined as:

$$v_i^E = 1 + E(d_i^E) \tag{8}$$

where $d_i^E$ is generated by $G_d(h_i)$ and means the probability of the final feature map ($h_i$) in image $i$ belonging to the source domain. $E(\cdot)$ means the information entropy that is the same as the one in formula (12). The more transferable the corresponding image is, the larger the entropy level attention value is. In the next section, we will introduce the minimum entropy regularization and how the entropy level attention affects the entropy loss.

### 3.4 Minimum entropy regularization (MER)

Minimum entropy regularization is designed to strengthen the prediction confidence of our model. Inspired by the idea of entropy function in information theory, entropy loss is proposed to reduce the uncertainty of probabilities for output classes. In 2005, Grandvalet and Bengio (2005) proposed the minimum entropy regularization (MER) for semi-supervised learning, and Long et al. (2016) firstly exploited the minimum entropy regularization on target domain data $D_T$, making the classifier more accessible to the unlabeled target data. In this section, we employ minimum entropy regularization for both target data and source data, and assign entropy level attention value to the entropy penalty. The benefits of this strategy are twofold. On the one hand, there exist great similarity between the oil palm tree type and the background type, which can easily be understood in Figure 1. MER helps to improve the prediction confidence for samples that are easy to be confused with other types. On the other hand, some images in the target domain are not transferable, such as the images with a low similarity in terms of the feature space across domains. Since these untransferable images are easier to be mistakenly classified, increasing their prediction confidence will confuse the classifier. To solve this problem, our entropy loss is weighted by the entropy level attention value, which is generated by a deep feature based adversarial discriminator. We embed the entropy level attention into the entropy loss according to formula (9):

$$L_E = -\frac{1}{n} \sum_{i=1}^{n} \sum_{c=1}^{C=4} v_i^E \cdot p_{i,c} \cdot log(p_{i,c}) \tag{9}$$

where $L_E$ is the entropy loss. $v_i^E$ is the entropy level attention. $C$ is the number of classes, which is 4 in our oil palm tree detection algorithm. $p_{i,c}$ is the prediction probability of classifier for image $x_i$ corresponding to class $c$, and we can acquire them according to the equation of $p_i = G_y(h_i)$, where $G_y$ is the classifier of our deep domain network and $h_i$ is attained from the transformed feature map $(H)$ with feature level attention.

In this way, the entropy level attention based minimum entropy regularization makes the prediction of our images more certain and confident and thus effectively improves the classifier's performance. Reasonably, our DA method with attention mechanism and minimum entropy regularization is naturally transferable across domains. In total, we can finally summarize our loss function as follows:

$$L = L_C^S + \mu * L_S + \alpha * L_D + \beta * L_E \tag{10}$$

where $\mu$, $\alpha$ and $\beta$ are the hyper-parameters that trade-off among shallow feature based domain loss, deep feature based domain loss and entropy loss. $L_C^S$ is the classification loss of the labelled source domain dataset and can be formulated as:

$$L_C^S = \frac{1}{n_S} \sum_{x_i \in D_S} L_y\left(G_y(h_i), y_i\right) \tag{11}$$

where $L_y$ is the cross-entropy loss function and $G_y$ is the classifier employed for making a final prediction for the source domain images.

**3.5 IOU based post-processing**

In previous sections, we discussed the training procedures of our proposed domain adaptive oil palm tree counting and detection, while in this part, we introduce our method in the prediction phase. First of all, we crop the original test image based on overlapping partitioning rules via a sliding window technique. We set the sliding step as 3 pixels following Li et al. (2017). After that, we predict each image using the MADAN model.

The right of Figure 4 illustrates the results after direct prediction, and we can see that there are many detected oil palms around one oil palm. Li et al. applied a time consuming method that iteratively merged the detected oil palms based on the distance (Li et al., 2016a). We adopt the IOU based principle to merge the detected oil palms that are close to each other. IOU is a popular evaluation metric used for measuring the accuracy of the detection results, and it is also used for merging detected objects in many end-to-end object detection algorithms. Here, we merge two detected oil palms if their IOU value is equal to or higher than a threshold and average their coordinates. So the final oil palm's coordinates can be calculated as:

$$(X_{lt}, Y_{lt}, X_{rb}, Y_{rb}) = \frac{1}{n} \sum_{i=1}^{n} (x_{lt,i}, y_{lt,i}, x_{rb,i}, y_{rb,i}) \tag{12}$$

where the subscripts $lt$ and $rb$ mean the left top and right bottom, $n$ represents the number of detected oil palms of which the IOU is lower than a threshold. IOU based merging is more efficient than distance-based merging as it only merges the detected oil palms once instead of iteratively. Ultimately, we accomplish the cross-regional oil palm tree detection via MADAN.

## 4. Study area and datasets

Table 1. The main information of Image A, Image B and Image C

| Index | Image A | Image B | Image C |
|---|---|---|---|
| Source | QuickBird | Google Earth | Google Earth |
| Longitude and latitude | 103.5991E, 1.5967N | 103.0518E, 5.0736N | 100.7772E, 4.1920N |
| Spectral | RGB, NIR | RGB | RGB |
| Acquisition date | November 21, 2006 | July 17, 2017 | December 21, 2015 |
| Resolution | 0.6 m | 0.3 m | 0.3 m |
| Image size | 12,188×12,576 pixels | 10,240×10,240 pixels | 10,496×10,240 pixels |
| Area | 55.18 km$^2$ | 9.44 km$^2$ | 9.67 km$^2$ |
| The number of oil palms | 291,827 | 47,917 | 91,357 |

Our study area locates in the Peninsular Malaysia (Figure 3), where the oil palm plantation is expanding increasingly and threatening the local environment and native species. According to the statistics in 2016, 47% of Malaysia oil palm plantation was in the Peninsular Malaysia (Tang et al., 2019). We have three high-resolution satellite images, Image A, Image B and Image C. Table 1 shows the elaborate information of these three satellite images. They are acquired from different sensors and locations, and the interval of photograph date is over 10 years, resulting in differences in reflectance, resolution, illumination and environmental conditions. As shown in Figure 4, samples of the same class in three images look quite distinguishing in characteristics and textures. Moreover, we preprocessed the three images in two steps. First, we employed a spectral sharpening method to Image A, and removed its NIR band. Second, to unify the resolution of these images to 0.6m, we downsampled Image B and Image C to $5,120 \times 5,120$ and $5,248 \times 5,120$ pixels by bilinear interpolation algorithm.

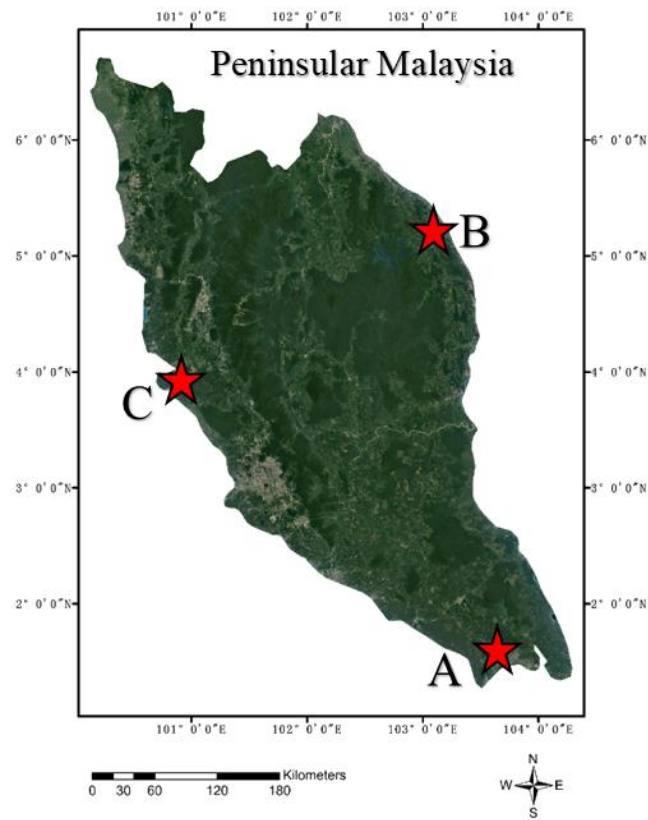

Figure 3. The location of our study area.

Figure 4 shows where and how our samples were collected. The training samples were collected from training areas (denoted by rectangles with solid lines) and validation samples were collected from the validation areas (denoted by rectangles with dotted lines). We manually interpreted four types of objects, including background, oil palm tree, other vegetation, and impervious or cloud. There are four types of objects in Image A and Image B, and three types of objects in Image C (without the type of impervious or cloud). The background and oil palm trees were collected from regions denoted by black rectangles. Other vegetation and impervious (or cloud) were collected from regions denoted by green squares and blue squares, respectively. We use 17 × 17 pixels as the sample size for all types following previous oil palm detection studies (Li et al., 2016; Li et al., 2019), which is similar to the largest crown size of a mature oil palm on these images. To evaluate our proposed MADAN for cross-regional oil palm tree detection, we detect the oil palms in the whole area of the three satellite images and compare the results with the manually annotated ground truth datasets.

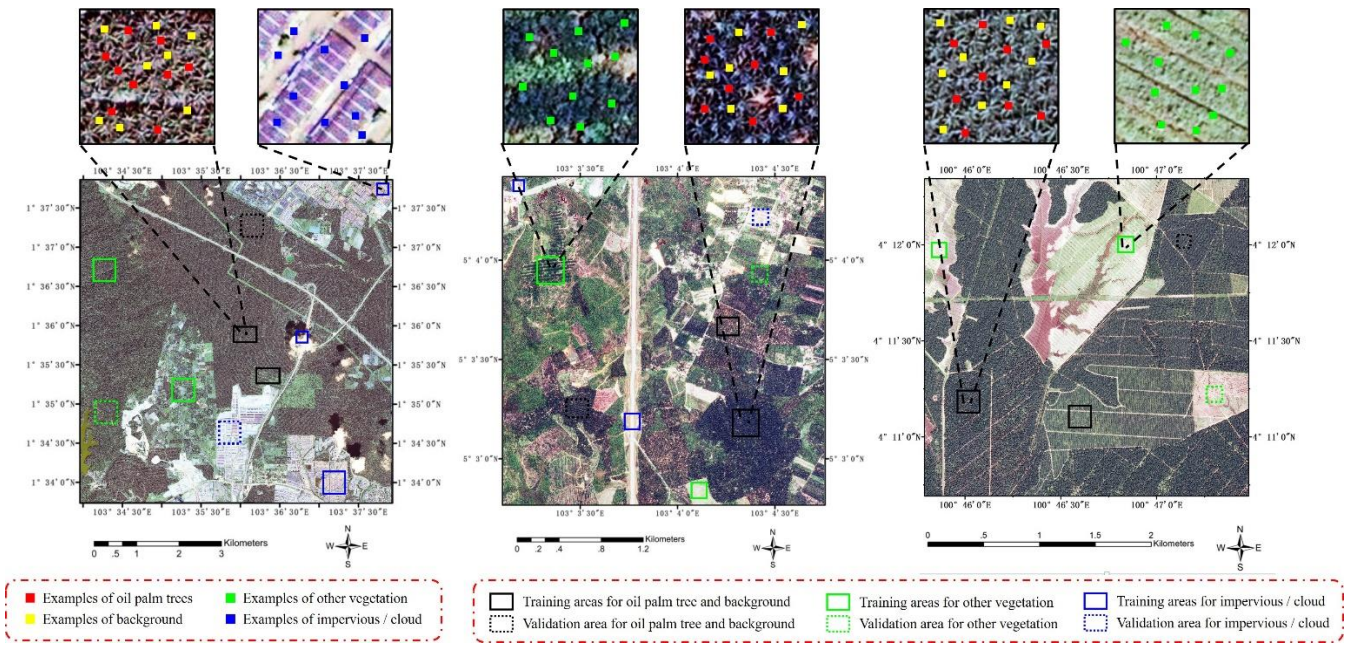

Figure 4. Examples of samples manually collected from different regions in Image A (left), Image B (middle) and Image C (right), respectively. Image A and Image B have four land cover types, while Image C has three land cover types (without impervious or cloud).

## 5. Experimental results

In this section, we evaluate the experiment results of our proposed MADAN for oil palm tree crown detection. First, we present the experimental setup and our evaluation metric in Section 5.1. And then we describe the oil palm tree detection results of MADAN in Section 5.2, followed by comparison with other state-of-the-art domain adaptation algorithms in Section 5.3.

## 5.1 Setup and evaluation metric

We complement our experiments based on the PyTorch deep learning framework (Paszke et al., 2017), and we set $\mu = \alpha = 0.1$ and $\beta = 1.0$ throughout all our experiments. The batch size is set as 128. The learning rate is 0.001. We train our model using GeForce RTX 2080 Ti. Moreover, we choose Adam (Kingma, & Ba, 2014) as our optimizer for training the domain discriminator and the classifier. We test our model after 20 epochs. The curve of training loss and the accuracy of validation dataset are illustrated in Appendix A.

Our evaluation metric consists of precision, recall and F1-score. Precision depicts the model's capability of detecting oil palms correctly, while recall describes the model's capability of detecting ground-truth oil palms. F1-score evaluates the overall performance of the model. They can be calculated from the following formulas:

$$precision = \frac{TP}{TP + FP} \times 100\% \qquad (13)$$

$$recall = \frac{TP}{TP + FN} \times 100\% \qquad (14)$$

$$F1 - score = \frac{2 * precision * recall}{precision + recall} \times 100\% \qquad (15)$$

where $TP$ means true positives, denoting the number of palms that are detected correctly; $FP$ means false positives, denoting the number of others that are detected as palms by mistake; $FN$ means false negatives, denoting the number of ground-truth palms that are missing in detection results. When the IOU metric value between the detected palm and a ground-truth oil palm tree is greater than or equal to 0.5, an oil palm tree will be called as correctly detected.

## 5.2 Oil palm tree counting and detection results via MADAN

To validate the performance of MADAN, we detect the oil palms across the whole satellite images. There are six transfer tasks: (1) Image A → Image B (A → B); (2) Image A → Image C (A → C); (3) Image B → Image A (B → C); (4) Image B → Image C (B → C); (5) Image C → Image A (C → A); (6) Image C → Image B (C → B). Table 3 displays the results of our proposed MADAN, with respect to TP, FP, FN, precision, recall, F1-score and average F1-score. We can find that our proposed method achieves an average F1-score of 84.81% for all six transfer tasks. In the meanwhile, when we set Image C as source domain, the detection results, especially the precision, is lower than other cross domain tasks. It might be the reason that Image C only has three object types (background, oil palm tree and other vegetation), causing more confusion between oil palms and other object types.

Table 3. The detection results of MADAN.

| Index | A → B | A → C | B → A | B → C | C → A | C → B |
|---|---|---|---|---|---|---|
| TP | 40,988 | 81,515 | 269,389 | 85,673 | 269,922 | 41,241 |
| FP | 8,048 | 3,830 | 63,420 | 15,414 | 105,944 | 19,498 |
| FN | 6,929 | 9,842 | 22,438 | 5,684 | 21,905 | 6,676 |
| Precision | 83.59% | 95.51% | 80.94% | 84.75% | 71.81% | 67.90% |
| Recall | 85.54% | 89.23% | 92.31% | 93.78% | 92.49% | 86.07% |
| F1-score | 84.55% | 92.26% | 86.25% | 89.04% | 80.85% | 75.91% |
| Average F1-score | 84.81% | | | | | |

## 5.3 Results comparison between MADAN and other DA approaches

We compare our proposed MADAN methods with other state-of-the-art DA methods. For traditional DA methods, we present the results of TCA (Pan et al., 2010) and GFK (Gong et al., 2012). As for deep learning-based DA methods, we select DDC (Tzeng et al., 2014), DAN (Long et al., 2015), DANN (Ganin et al., 2016) and Deep CORAL (Sun, & Saenko, 2016). Table 4 lists the results of above mentioned DA methods and our MADAN methods. We also show the F1-score of the Baseline method (AlexNet based method trained by the labeled source dataset) and the upper bound (AlexNet based method trained by the labeled target dataset) as a reference. We illustrate the detection results in 24 regions for 6 transfer tasks (4 regions for each task). We can observe the detection results in Figure 5-10, which describe the performance of one example region for 6 transfer tasks. More detection results can be found in Appendix B. The green points denote the correct detected oil palms, the yellow circles denote the ground-truth oil palms that are missing, and the red squares with red points denote other types of objects like other vegetation or building corners that are detected as oil palms by mistaken. Results demonstrate that MADAN outperforms other DA methods in all six transfer tasks except Image B → Image C, and achieves the highest average F1-score among eight methods The F1-score of MADAN is very close to the upper bound for A → B and A → C. We also evaluate the impact of data augmentation (horizontal flipping, vertical flipping and brightness transformation) on the Baseline and MADAN methods in Table 5. Experimental results show that the data augmentation strategy improves the detection accuracy of baseline method by 2.04%, while has little impact on the results of MADAN. Furthermore, we list the efficiency of different DA methods in Table 6. Although our method has the largest number of parameters and the FLOPs (floating point of operations), the inference time (ms per image) is comparable with other DA methods. TCA and GFK are quite slow due to the complicated matrix transformation and iteration in computation progress.

Table 4. The F1-scores of different DA methods for all six transfer tasks.

| Index | A → B | A → C | B → A | B → C | C → A | C → B | Average |
|---|---|---|---|---|---|---|---|
| Baseline | 65.63% | 81.38% | 70.15% | 77.39% | 62.74% | 55.39% | 68.78% |
| TCA | 64.36% | 85.21% | 66.47% | 76.71% | 65.84% | 63.30% | 70.32% |
| GFK | 66.49% | 88.55% | 69.34% | 78.21% | 69.70% | 63.09% | 72.56% |
| DANN | 67.05% | 85.84% | 71.82% | 81.76% | 70.06% | 64.09% | 73.44% |
| DDC | 80.92% | 89.39% | 73.87% | 86.14% | 71.45% | 63.40% | 77.53% |
| Deep CORAL | 78.92% | 86.79% | 74.96% | 89.35% | 72.30% | 66.10% | 78.07% |
| DAN | 77.90% | 88.54% | 80.24% | **91.62%** | 75.91% | 73.32% | 81.26% |
| MADAN (ours) | **84.55%** | **92.26%** | **86.25%** | 89.04% | **80.85%** | **75.91%** | **84.81%** |
| Upper bound | 85.70% | 94.02% | 90.68% | 94.02% | 90.68% | 85.70% | 90.13% |

Table 5. The F1-score of Baseline and MADAN with/without augmentation strategy.

| Index | A → B | A → C | B → A | B → C | C → A | C → B | Average |
|---|---|---|---|---|---|---|---|
| Baseline | 65.63% | 81.38% | 70.15% | 77.39% | 62.74% | 55.39% | 68.78% |
| Baseline + Augmentation | 68.04% | 83.06% | 71.05% | 83.68% | 64.59% | 54.48% | 70.82% |
| MADAN | 84.55% | 92.26% | 86.25% | 89.04% | 80.85% | 75.91% | 84.81% |
| MADAN + Augmentation | 83.18% | 93.15% | 82.02% | 88.64% | 82.49% | 75.02% | 84.08% |

Table 6. The efficiency of different DA methods.

| Index | Number of Parameters (M) | GFLOPs | Inference time (ms per image) |
|---|---|---|---|
| Baseline | 56.82 | 0.43 | 1.54 |
| TCA | 2.25 | 0.32 | 8.66 |
| GFK | 2.25 | 0.32 | 10.15 |
| DANN | 67.31 | 0.45 | 1.80 |
| DDC | 56.82 | 0.43 | 1.76 |
| Deep CORAL | 56.82 | 0.43 | 1.72 |
| DAN | 56.82 | 0.43 | 2.04 |
| MADAN (ours) | 77.80 | 0.48 | 2.01 |

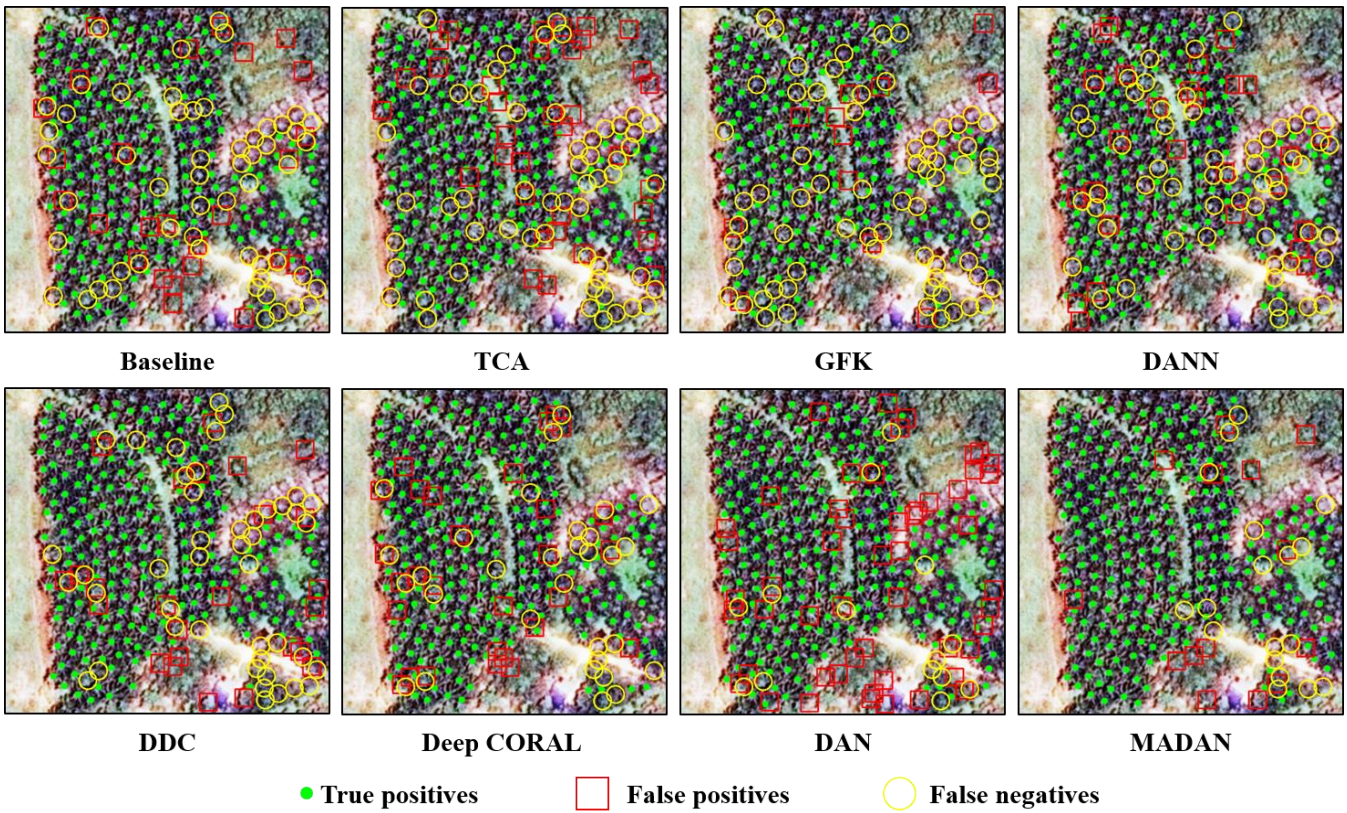

Figure 5. The detection results in Region 1 for Image A → Image B.

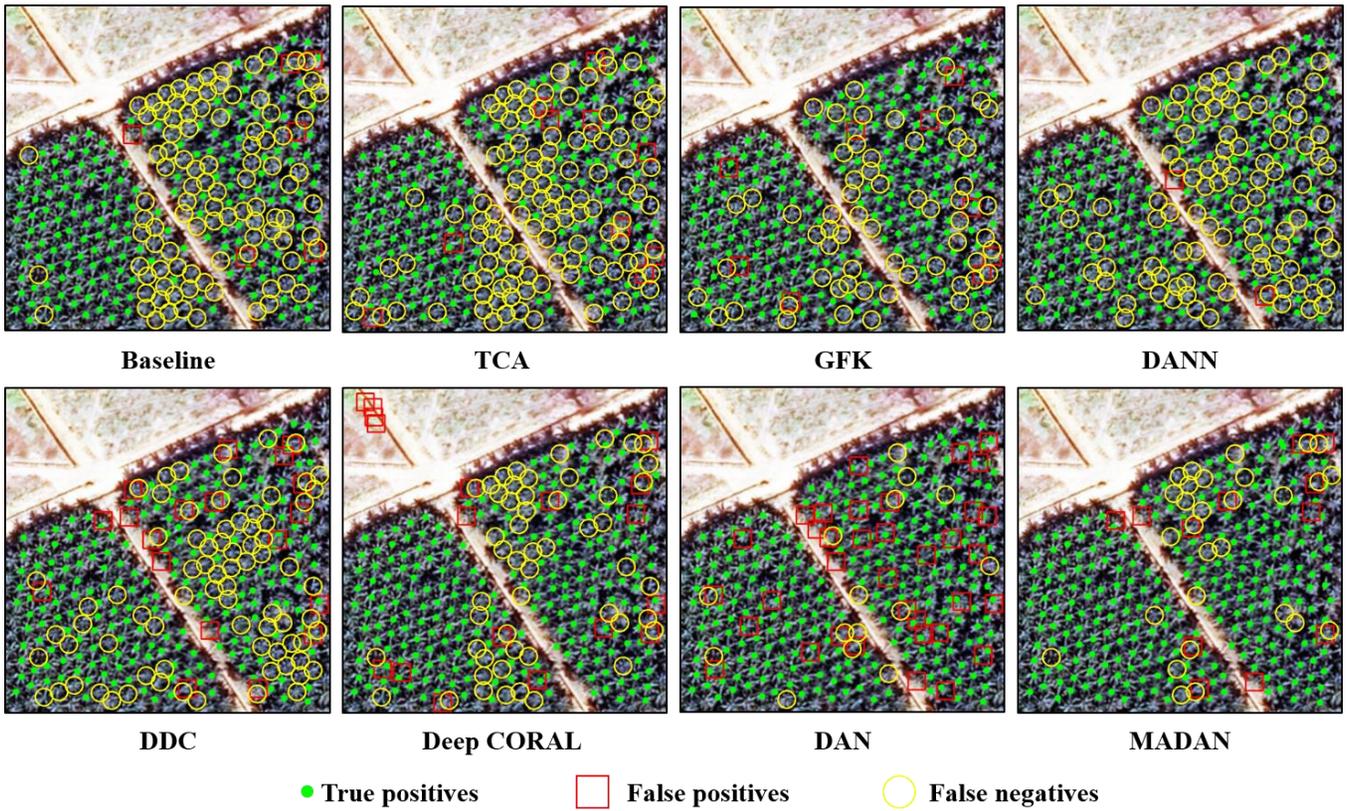

Figure 6. The detection results in Region 1 for Image A → Image C.

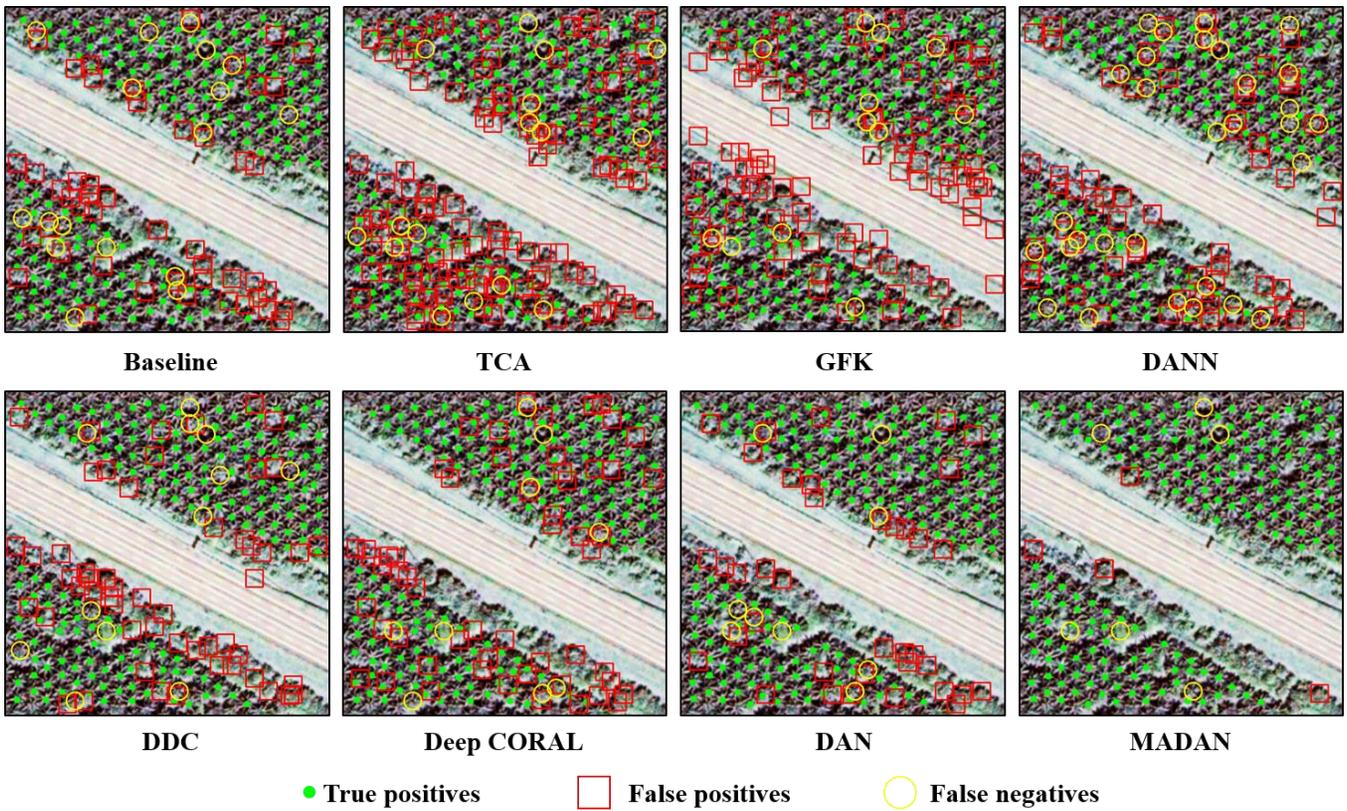

Figure 7. The detection results in Region 1 for Image B → Image A.

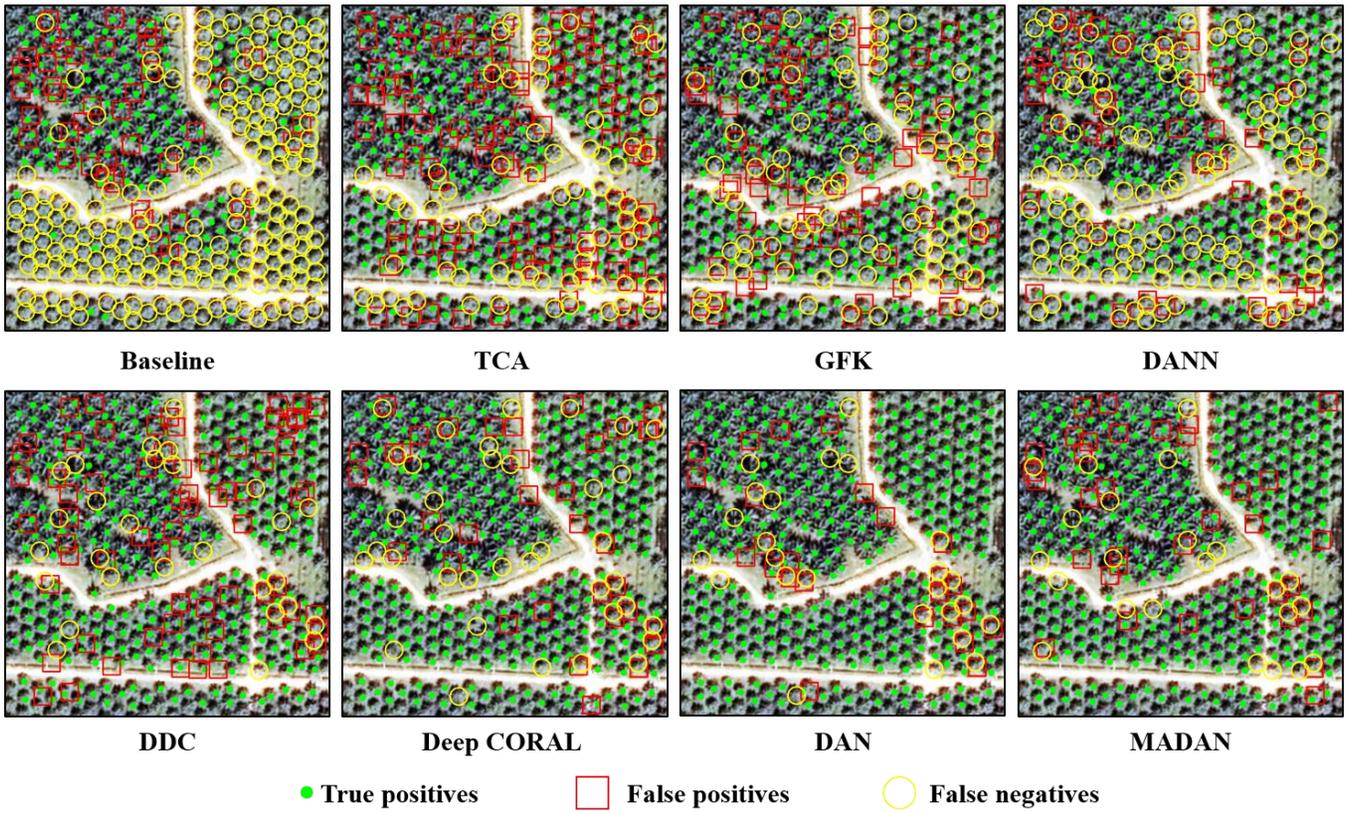

Figure 8. The detection results in Region 1 for Image B → Image C.

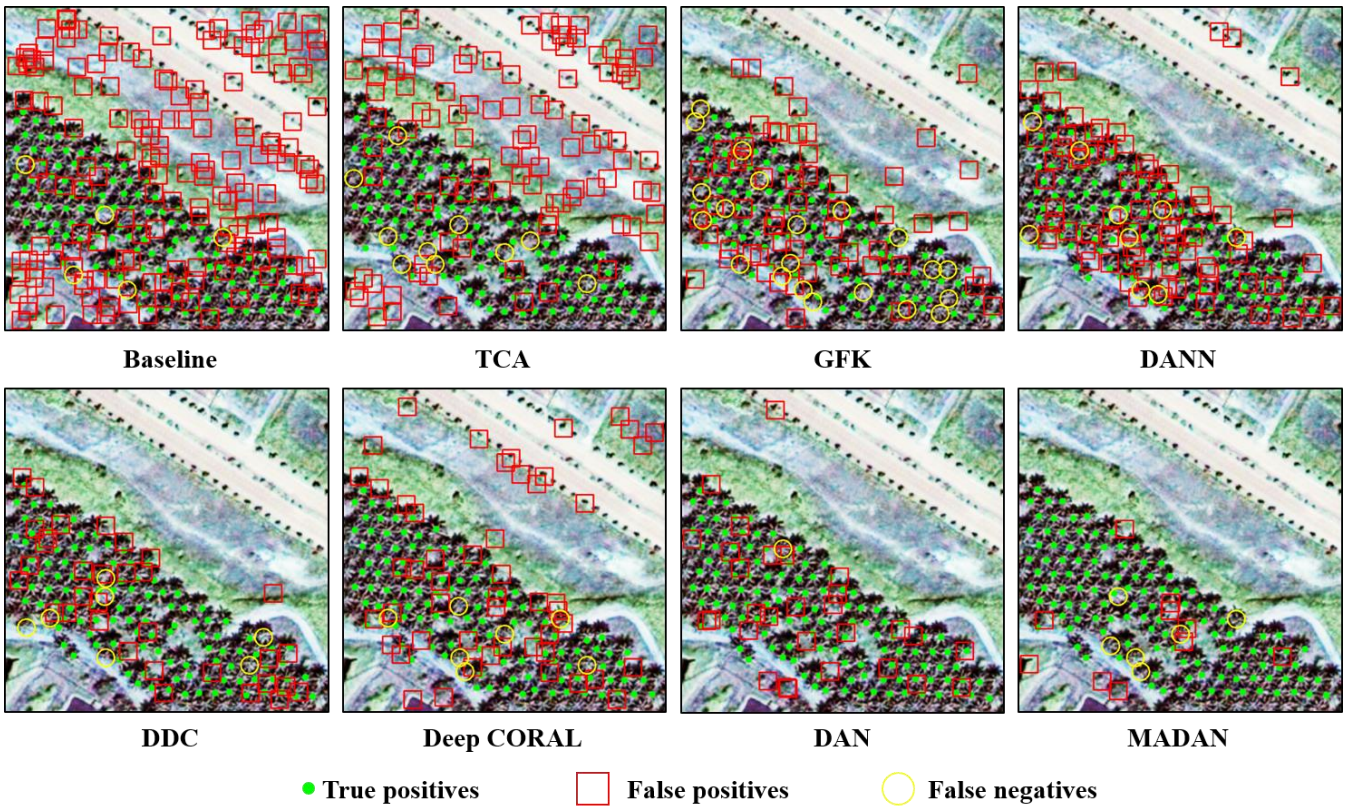

Figure 9. The detection results in Region 1 for Image C → Image A.

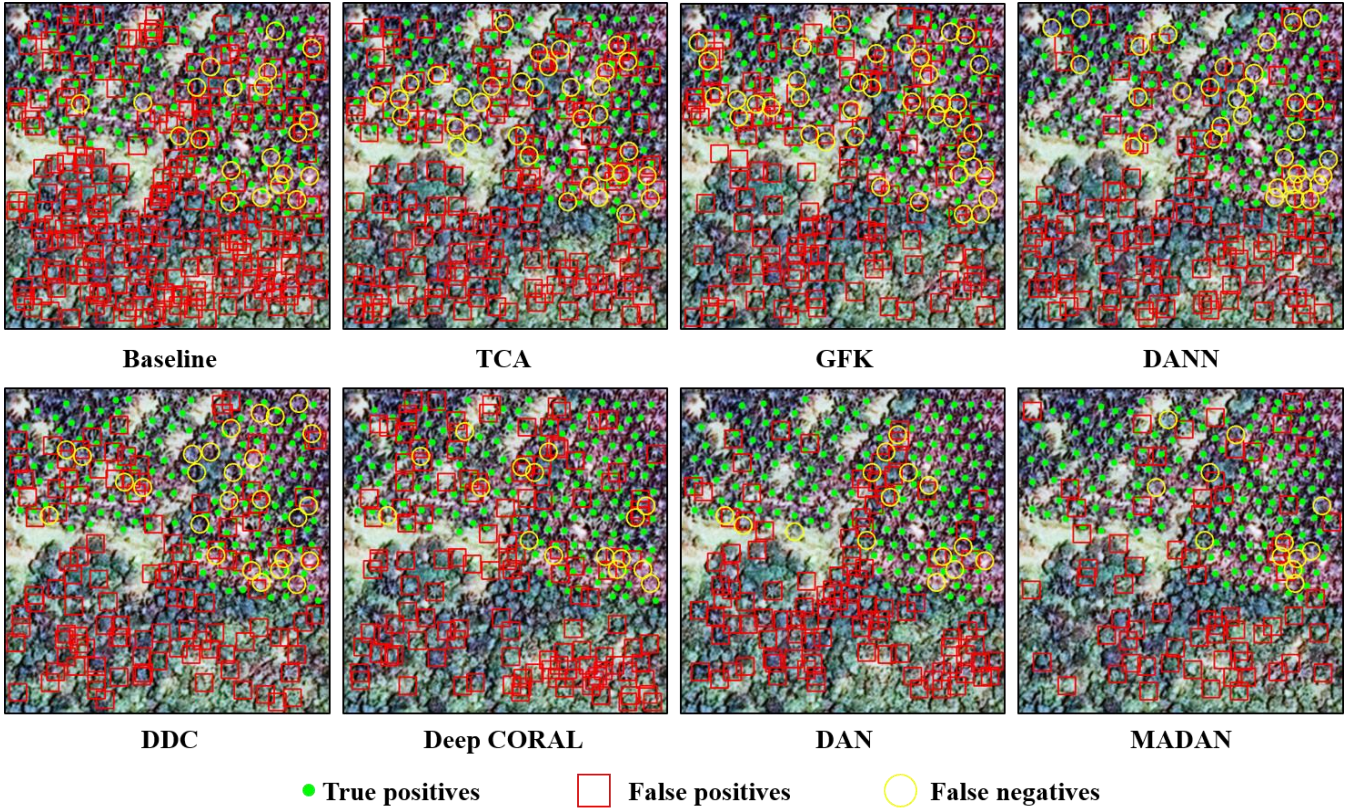

Figure 10. The detection results in Region 1 for Image C → Image B.

## 6. Discussion

In this part, we assess the effectiveness of each strategy in our proposed MADAN through comprehensive ablation experiments. Table 7 shows the results of Baseline, Baseline + BIN, MADAN (without MER), and MADAN. We present a comprehensive analysis of each strategy as follows. Notably, in BIN, we only employ the labeled source domain dataset without the target domain dataset. In MLA and MER, we employ both the labeled source domain dataset and the unlabeled target domain dataset.

Table 7. The F1-scores of Baseline, Baseline + BIN, MADAN (without MER) and MADAN.

| Index | A → B | A → C | B → A | B → C | C → A | C → B | Average |
|---|---|---|---|---|---|---|---|
| Baseline | 65.63% | 87.70% | 70.15% | 77.39% | 62.74% | 55.39% | 69.93% |
| Baseline + BIN | 68.67% | 86.25% | 73.94% | 80.68% | 68.30% | 59.43% | 72.88% |
| MADAN (without MER) | 81.36% | 89.90% | 80.28% | 84.43% | 76.24% | 68.37% | 80.10% |
| **MADAN** | **84.55%** | **92.26%** | **86.25%** | **89.04%** | **80.85%** | **75.91%** | **84.81%** |

## 6.1 Ablation study of the BIN block.

To explicitly explore how BIN achieves better generalization, we analyze the feature divergence caused by domain bias. In this paper, we select the features in ReLU layer (denoted by $R$) to calculate the divergence between the source and the target datasets. Following previous studies (Li et al., 2016a; Tu et al., 2019), we assume a Gaussian distribution of $R$, with mean $\mu$ and variance $\sigma^2$. Our divergence can be calculated as follows:

$$D(R_S \parallel R_T) = KL(R_S \parallel R_T) + KL(R_T \parallel R_S) \tag{16}$$

$$KL(R_S \parallel R_T) = \log\frac{\sigma_S}{\sigma_T} + \frac{\sigma_S^2 + (\mu_S - \mu_T)^2}{2\mu_T^2} - \frac{1}{2} \tag{17}$$

where $R_S$ and $R_T$ denote the features from the source dataset and the target dataset, respectively. $KL(R_S \parallel R_T)$ means Kullback-Leibler (KL) divergence between the source feature and the target feature. $D(R_S \parallel R_T)$ means the symmetric KL divergence between $R_S$ and $R_T$. The average divergence of all layers can be formulated as:

$$D(L_S \parallel L_T) = \frac{1}{L * C} \sum_{l=1}^{L} \sum_{c=1}^{C} D(R_S^{l,c} \parallel R_T^{l,c}) \tag{18}$$

where $R_S^{l,c}$ represents the source feature of the $c^{th}$ channel in the $l^{th}$ layer. $L$ and $C$ means the number of layers in the network and the number of channels in a certain layer, respectively (Li et al., 2016). The smaller the average divergence is, the more powerful the generalization of network is.

We evaluate the mean feature divergence of Baseline and Baseline + BIN for all six transfer tasks in Figure 11 and our BIN blocks obviously reduce the feature divergence between source and target domains. The final detection results listed in Table 5 are consistent with the performance of feature divergence. Results demonstrate that our BIN based feature extractor can effectively enhance the generalization ability of the model and improving the accuracy of the "unseen" target domain dataset when only using training samples of the source domain.

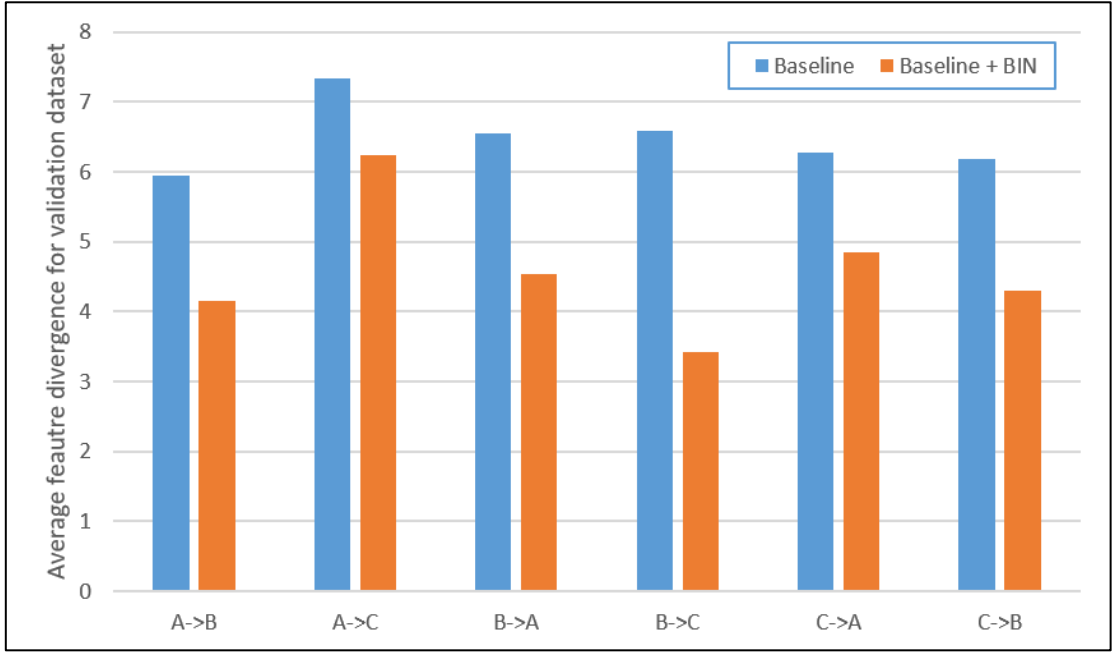

Figure 11. The average feature divergence of all ReLU layers among different domains for Baseline and Baseline + BIN. The final detection results are consistent with the performance of feature divergence.

**6.2 Ablation study of the minimum entropy regularization.**

In section 4.4, we propose an entropy loss weighted by entropy level attention value for MER, Table 5 shows that MER improves the F1-scores by 2.36-7.54% for six transfer tasks, and improves the average F1-score by 4.71%. Figure 12 further demonstrates the effectiveness of MER. The x-axis denotes 4 classes and the y-axis denotes the prediction probabilities of them. After embedding MER, the distribution of the prediction probabilities is unimodal instead of bimodal. We can conclude that our proposed MER enables the model to make predictions more confidently.

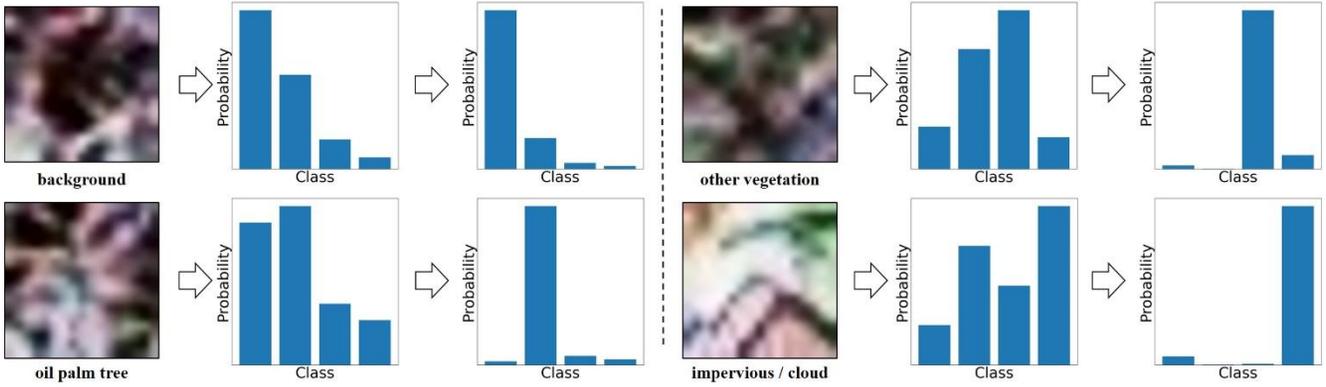

Figure 12. The effectiveness of MER. These four samples are from Image B → Image A task. For each group, the left image is the sample image, while the middle and the right histogram is the prediction probabilities of MADAN without and with MER, respectively. The x-axis denotes 4 classes including background, oil palm tree, other vegetation and pervious / cloud. The y-axis denotes the prediction probabilities of above 4 classes. We can find that the entropy loss weighted by entropy level attention value enables the model to make predictions more confidently.

**6.3 Ablation study of the multi-level attention mechanism.**

In this section, we evaluate the performance of the MLA. Table 8 lists the results of ablation experiments of the multi-level attention mechanism. Based on the 4 types of loss including shallow feature based domain loss, deep feature based domain loss, classifier loss and entropy loss, we compare the following four cases to evaluate the separate contributions of the feature level attention and the entropy level attention: (1) no attention mechanism (denoted by Baseline + no attention); (2) only feature level attention mechanism (denoted by Baseline + feature level attention); (3) only entropy level attention mechanism (denoted by Baseline + entropy level attention); (4) the combination of feature level attention and entropy level attention (denoted by Baseline + multi-level attention). Similarly, we compare the above four cases based on BIN extractor (denoted by BIN + no attention, BIN + feature level attention, BIN + entropy level attention, and MADAN, respectively). Results show that our proposed MLA improves the average F1-score of Baseline method by 4.41%, and improves the average F1-score of Baseline + BIN method by 5.25%. The entropy level attention performs better than the feature level attention, and integrating both of them (MLA) obtains the highest average F1-score. Additionally, we display the attention values of several samples in Figure 13. We can observe that the samples with more transferable context often have a higher attention value.

Table 8. The F1-scores of ablation experiments about multi-level attention mechanism

| Index | A → B | A → C | B → A | B → C | C → A | C → B | Average |
|---|---|---|---|---|---|---|---|
| Baseline + no attention | 72.83% | 87.97% | 74.53% | 84.06% | 72.63% | 65.66% | 76.28% |
| Baseline + feature level attention | 76.05% | 88.79% | 74.16% | 87.93% | 72.70% | 65.75% | 77.56% |
| Baseline + entropy level attention | 77.47% | 89.90% | 79.32% | 88.46% | 73.35% | 63.26% | 78.63% |
| Baseline + multi-level attention | 80.22% | 91.67% | 80.83% | 89.97% | 75.24% | 66.79% | 80.79% |
| BIN + no attention | 79.23% | 88.54% | 80.11% | 88.18% | 74.28% | 66.99% | 79.56% |
| BIN + feature level attention | 81.18% | 90.06% | 83.45% | 90.08% | 76.45% | 67.88% | 81.52% |
| BIN + entropy level attention | 83.23% | **92.30%** | 86.04% | 88.95% | **81.18%** | 70.75% | 83.74% |
| BIN + multi-level attention (MADAN) | **84.55%** | 92.26% | **86.25%** | **89.04%** | 80.85% | **75.91%** | **84.81%** |

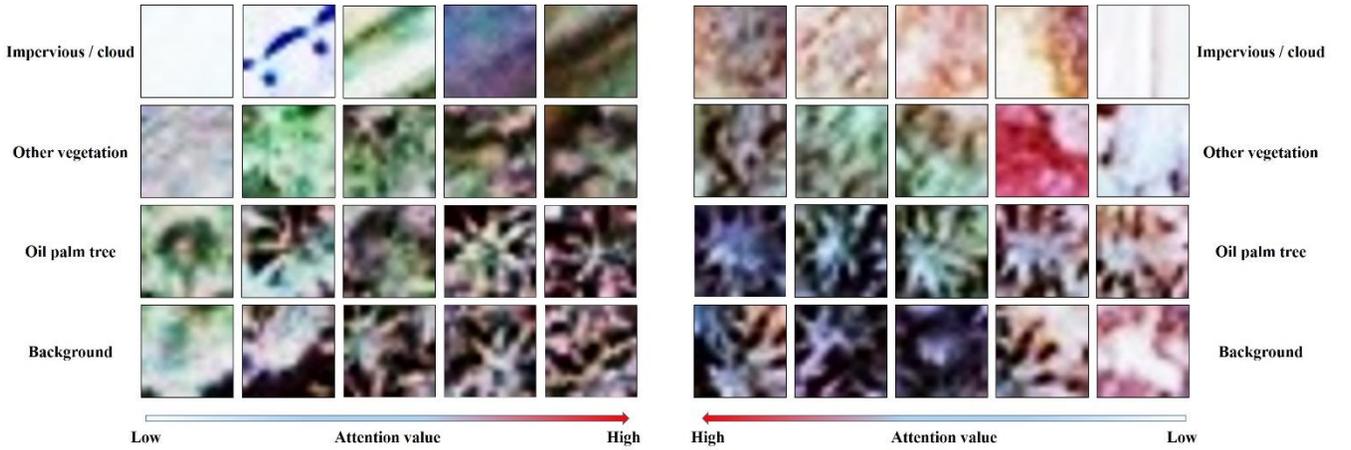

Figure 13. The feature level attention value of some target domain samples for Image B → Image A task (left) and Image A → Image B task (right). The samples with more transferable context often have a higher attention value.

## 7. Conclusions

In this paper, we propose a novel domain adaptive oil palm tree counting and detection method, i.e., a Multi-level Attention Domain Adaptation Network (MADAN). MADAN comprises four procedures: BIN based feature extractor, multi-level attention mechanism (MLA), minimum entropy regularization (MER), and IOU based post-processing. We integrate instance normalization and batch normalization into our BIN block, which effectively enhances the generalization performance of our network only with source domain dataset. Our proposed multi-level attention mechanism is generated by an adversarial neural network, including feature level attention and entropy level attention. Feature level attention applies weights to the final feature map and entropy level attention applies weights to the entropy of label prediction. MLA improves the transferability of model with unlabeled target domain images. Furthermore, we present minimum entropy regularization with entropy loss to make our prediction more confident. As for reference phase, we adopt a sliding window based technique and an IOU based post-processing to acquire final oil palm detection results for the target image.

We evaluate our proposed method using three large-scale satellite images (denoted by Image A, Image B and Image C) located in the Peninsular Malaysia. Our comprehensive ablation experiments show that our BIN based extractor and multi-level attention mechanism increase the capacity of generalization and transferability, respectively. Only with labelled source domain images, BIN based feature extractor improves the average F1-score by 3.05% compared with Baseline. After adding unlabeled target domain images, MLA increases the average F1-score by 11.93% compared with Baseline + BIN. MER enables our cross-regional oil palm detection model to make predictions more confidently, improving the average F1-score by 3.29% compared with Baseline + BIN + MLA. Based on the above three strategies, our proposed MADAN improves the average F1-score by 14.98% for all six transfer tasks, compared with the Baseline method without using DA approach. MADAN achieves an average F1-score of 84.81% without any target domain annotation, which are very close to the upper bound (trained by labeled target datasets) for several transfer tasks. Our MADAN outperforms other existing domain adaptation methods like DAN, DDC, Deep CORAL, etc., improving the F1-scores by 3.55%-14.49%. In the future, we will explore and develop more effective DA algorithms, and apply them to end-to-end oil palm counting and detection methods. We will also detect oil palm trees in a larger-scale and more complex area using multi-source and multi-temporal remote sensing images.

**Acknowledgements**

This research was supported in part by the National Key Research and Development Plan of China (Grant No. 2017YFA0604500, 2017YFB0202204 and No.2017YFA0604401), the National Natural Science Foundation of China (Grant No. 51761135015), and by Center for High Performance Computing and System Simulation, Pilot National Laboratory for Marine Science and Technology (Qingdao). In addition, we would like to thank Mr. Ximei Wang for his valuable discussion.